\documentclass[11pt]{article}
\usepackage{bm, amsmath, amsfonts, amssymb,lscape}
\usepackage{graphicx}
\usepackage{amssymb}
\usepackage{bm}                
\usepackage{mathrsfs}      
\usepackage{amsmath, amsthm}
\usepackage{graphicx,epsf,subfigure,pstricks,pst-node,psfrag,amsthm,amssymb,amsmath}
\usepackage{float}
\usepackage{array}
\usepackage{hyperref}
\usepackage{rotating}
\usepackage{natbib}

\usepackage{graphicx}
\usepackage{amssymb}
\usepackage{bm}                
\usepackage{mathrsfs}      
\usepackage{amsmath, amsthm}
\usepackage{graphicx,epsf,subfigure,pstricks,pst-node,psfrag,amsthm,amssymb,amsmath}
\usepackage{float}
\usepackage{array}
\usepackage{hyperref}
\usepackage{authblk}
\usepackage{natbib}

\numberwithin{equation}{section}

\usepackage{JASA_manu}

\begin{document}

\title{Elliptical modeling and pattern analysis for perturbation models and classification}
\author[1]{Shan Suthaharan}
\author[2]{Weining Shen}
\affil[1]{Department of Computer Science, University of North Carolina at Greensboro, Greensboro}
\affil[2]{Department of Statistics, University of California, Irvine}
\date{}
\maketitle

\begin{center}
\textbf{Abstract}
\end{center}
The characteristics (or numerical patterns) of a feature vector in the transform domain of a perturbation model differ significantly from those of its corresponding feature vector in the input domain. These differences - caused by the perturbation techniques used for the transformation of feature patterns - degrade the performance of machine learning techniques in the transform domain. In this paper, we proposed a nonlinear parametric perturbation model that transforms the input feature patterns to a set of elliptical patterns, and studied the performance degradation issues associated with random forest classification technique using both the input and transform domain features. Compared with the linear transformation such as Principal Component Analysis (PCA),  the proposed method requires less statistical assumptions and is highly suitable for the applications such as data privacy and security due to the difficulty of inverting the elliptical patterns from the transform domain to the input domain. In addition, we adopted a flexible block-wise dimensionality reduction step in the proposed method to accommodate the possible high-dimensional data in modern applications. We evaluated the empirical performance of the proposed method on a network intrusion data set and a biological data set, and compared the results with PCA in terms of classification performance and data privacy protection (measured by the blind source separation attack and signal interference ratio). Both results confirmed the superior performance of the proposed elliptical transformation.   


\newpage

\newtheorem{theorem}{Theorem}
\newtheorem{lemma}{Lemma}
\newtheorem{remark}{Remark}
\newtheorem{definition}{Definition}
\newtheorem{example}{Example}
\newtheorem{corollary}{Corollary}
\newtheorem{ex}{}
\newtheorem{proposition}{Proposition}

\section{Introduction}
Feature vectors carry useful numerical patterns that characterize the original domain (or a sub original domain - input domain) formed by the feature vectors themselves. Machine learning algorithms generally utilize these patterns to generate classifiers, that can help make decisions from data, by using supervised or unsupervised learning techniques \citep{suthaharan2015machine}. However, certain data science applications, such as data privacy and data security \citep{whitworth2014security}, require the alteration of these feature patterns to protect data privacy so that it should be difficult to recover the original patterns from the altered patterns \citep{little1993statistical}. Perturbation models have been studied and developed for this purpose \citep{muralidhar2003theoretical} and \citep{FienbergSteele1998}. The perturbation models generally transform the feature vectors from an original domain to a new set of feature vectors within a transform domain where the data privacy can be protected. On the other hand, the performance of machine learning algorithms can be degraded in the transform domain due to the alternations of the patterns. Hence a significant research has been performed to develop an efficient perturbation model to minimize the degradation of the performance of machine learning algorithms while providing a robust protection of data privacy. Perturbation models may be categorized into two top-level groups: parametric models and nonparametric models. The parametric models may also be further divided into two subgroups: vector space (or the original domain) models and feature space (or the transform domain) models. The vector space models include the models proposed by \citep{muralidhar2003theoretical}, in which the authors have shown that their proposed models perform well in the original domain. Alternatively,  \citep{oliveira2004achieving} proposed a feature space model which was constructed using a matrix rotation, and \citep{lasko2010spectral} also developed a feature space model, but they used a spectral analysis. They showed their proposed techniques performed well in the transform domain. These types of models make parametric statistical assumption which in practice can be easily violated for different types of data. As a consequence, the current techniques may not perform as desired. A thorough review was presented in a recent paper by \citep{qian2015drive}, in which the authors summarized the possible types of violations of parametric assumptions, including uncertainty in marginal distributional properties of independent variables and possible nonlinear relationship that linear models cannot fully explore (e.g., invert-U shape \citep{aghion2005competition}). They proposed a nonparametric model based on density ratios to address  these problems and reported that the nonparametric models in general can perform better than the other parametric models. 

In this paper, we also considered a nonparametric perturbation model without imposing any parametric assumptions on the marginal distribution of features. The main idea is to construct a transform domain (or feature space) from the original domain using parametrized elliptical patterns with the goals of making the restoration of the original patterns very difficult, while maintaining a similar performance for the machine learning algorithms in both the original and the transform domains. Our proposed approach, Elliptical Pattern Analysis (EPA),  sets the criteria on privacy strength based on blind source separation attack \citep{zarzoso1999blind}, because of the use of mutual interaction between variables to construct transform domain. 

\begin{figure}[!t]
\centering
\includegraphics[scale=.6]{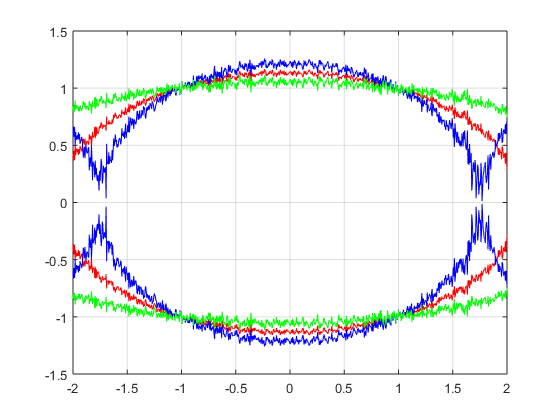}
\caption{Three ellipses generated by equation (4) using 3 sets of parameter values: (0.22, 0.78, 0.1) - highlighted using Red color; (0.32, 0.68,0.04) - highlighted using Blue color; (0.1, 0.9, 0.05) - highlighted using Green color for the parameters (a, b, $\alpha$), respectively. It shows some signal interference between the elliptical patterns distorted by the noise parameter $\alpha$.}
\label{fig:ell1}
\end{figure}

Our key contribution includes the use of mutual interaction between two variables (or features); however, this type of aggregation may jeopardize the performance of classification algorithms through the loss of some of the data characteristics (or patterns). To solve this problem, we proposed an additional data aggregation step through the random projection in the feature space before applying any machine learning algorithms. The main idea is to search over possible ways to combine pairs (or blocks) of variables to achieve efficient dimension reduction while maintaining useful predictive information to help later-stage for machine learning algorithms. In particular, we consider classification algorithms and use random forest classification on the reduced feature space. By aggregating feature variables, the proposed method significantly enhances the protection of data privacy and reduces computational cost. 
\begin{figure}[!t]
\centering
\includegraphics[scale=.6]{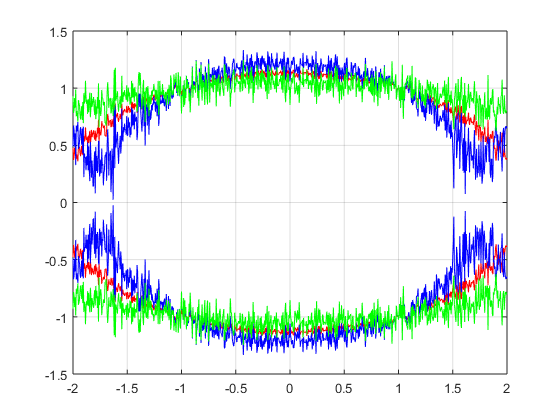}
\caption{Three ellipses generated by equation (4) using 3 sets of parameter values: (0.22, 0.78, 0.05) - highlighted using Red color; (0.32, 0.68,0.10) - highlighted using Blue color; (0.1, 0.9, 0.15) - highlighted using Green color for the parameters (a, b, $\alpha$), respectively. It illustrates a significant signal interference between the ellipses distorted by a very high noise.}
\label{fig:ell2}
\end{figure}

\section{A Perturbation Model}
We define the proposed EPA approach as a model that transforms a sub original domain (input domain) through a perturbation process such that the feature vector is altered in the transform domain to achieve a set of specific recommended goals - the goals that lead to the protection of data privacy and the generation of classifiers. In this section, the perturbation models is defined using a mathematical transformation ($T$) and recommended quantitative measures for quantifying the strength of data privacy ($\rho$) and misclassification error ($\eta$ or $\zeta$). 

\subsection{Mathematical Definition}
Suppose $\textbf{x}$ is a feature vector with dimension $p$ in the input domain $X$, and $\textbf{y}$ is its perturbed feature vector with dimension $q$ (where $q < p$) in the transform domain $Y$, then we define the mathematical relationship between $\textbf{x}$ and $\textbf{y}$ as follows:
\begin{equation}
\textbf{y} = T(\textbf{x}),
\end{equation}
where the mathematical transformation $T$ defines the proposed perturbation model, and its intention is to satisfy the condition $\rho(\textbf{x},\textbf{y})> \epsilon_0$ for some quantitative measure $\rho$. In other words, this condition describes the difficulty of recovering the feature vector $\textbf{x}$ from the feature vector $\textbf{y}$ given the transformation $T$ and the quantitative measure $\rho$. One of the applications that satisfy this type of modeling is data privacy where the owner of the data wants to share the data to an intended user, while its privacy is protected, given the transformation $T$ and the measure $\rho$ are chosen appropriately.

\subsection{Problem Definition}
The condition imposed on the proposed perturbation model can adversely affect other applications that require the use of a feature vector in the transform domain to achieve similar or better classification results obtained with the feature vector of the input domain, along with data privacy. Suppose $\eta$ is a performance measure (e.g. misclassification error) of an application $M$, then the performance degradation of the perturbation model $T$ can be defined as follows: 
\begin{equation}
\eta (M(\textbf{y})) > \eta (M(\textbf{x})),
\end{equation}
where $\textbf{y}=T(\textbf{x})$ and we define the degradation measure as follows: $\zeta_T(\textbf{x},\textbf{y})=\eta(M(\textbf{x})) - \eta(M(\textbf{y}))$. While it is expected that $\zeta_T(\textbf{x},\textbf{y}) \leq 0$ for a perturbation model, it is also possible that we get $\zeta_T(\textbf{x},\textbf{y}) > 0$; that is better performance with $\textbf{y}$ for a perturbation model. The application $M$ that we consider in this paper is a classification technique - in particular the random forest technique - with the misclassification error as the performance measure $\eta$.


\section{The Proposed Methodology}
This study requires - as per the definitions and problems stated in the previous section - a perturbation model $T$ with its condition measure $\rho$, and an application $M$ with its performance degradation measure $\eta$. They are presented in this section with a detailed discussion.

\subsection{Elliptical Perturbation Model}
Our feature vector $\textbf{x}$ in the input domain may be represented by $p$ variables (or features), $x_1, x_2, \dots, x_p \geq 0$. We also assume $p$ is an even integer without loss of generality. We use the proposed perturbation on consecutive pairs of variables: $(x_1,x_2)$, $(x_3,x_4)$, $\dots$, $(x_{p-1},x_p)$ to generate the feature vector $\textbf{y}$ which is represented by new variables $y_1,y_2,\ldots,y_q$; $q=p/2$, respectively. Take $(x_1,x_2)$ as an example, we consider
\begin{equation}
y_1  = \sqrt{a x_1^2 + b x_2^2} + \alpha \epsilon,
\label{eq:three}
\end{equation}
where $a$ and $b$ are unknown parameters, $\epsilon \sim N(0, 1)$ and $\alpha$ determines the strength of noise degradation. To further simplify the process, we can assume $a+b=1$ and $a,b \geq 0$. The model reduces to the standard linear model when $a \rightarrow 0$ or $a \rightarrow 1$. The nonlinear transformation $\sqrt{a x_1^2 + bx_2^2}$ defines the elliptical perturbation model and describes the nonlinear mutual interaction between the feature variables $x_1$ and $x_2$. 

On one hand, we can choose the value for $a$ such that the classification results using $\textbf{y}$ and $\textbf{x}$ are significantly close to each other (i.e., $\zeta_T \sim 0$). On the other hand, we can choose $a$ to minimize the absolute value of correlations between $y_1$ and $(x_1,x_2)$. Meanwhile, noise strength $\alpha$ will be tuned to achieve the intended goal (e.g. data privacy determined by $\rho$) of the perturbation model. In the model building process, we will use this correlation-minimization to tune the parameter $a$.

\subsection{Elliptical Patterns Visualization}

The visual interpretation of the studied model in equation (\ref{eq:three}) is presented in Fig. \ref{fig:ell1}. We have illustrated the elliptical characteristics of the model by fixing the variable $y$ to a single value and varying the values of the parameters $a$, $b$, and $\alpha$. For simplicity, we have selected $y=1$, and a set of values (0.22, 0.78, 0.03), (0.32, 0.68, 0.04), and (0.1, 0.9, 0.05) for the parameters $a$, $b$, $\alpha$, respectively. The model in equation (\ref{eq:three}), with these values, provides the three elliptical patterns with interference characteristics as illustrated in Fig. \ref{fig:ell1}. In order to generate these elliptical patterns, we transform equation (\ref{eq:three}) as follows:
\begin{equation}
x_2 =  \sqrt{\frac{{(y_1-\alpha \epsilon)}^2 + a x_1^2}{b}},
\label{eq:three-b}
\end{equation} 
It clearly shows the difficulty of finding a pair of $(x_1, x_2)$ for a given value of $y_1$ under a scaled noise degradation due to elliptical interference. To illustrate the strength of the model visually, we increased the values of $\alpha$ from 0.03, 0.04, and 0.05 to 0.05, 0.1, and 0.15, respectively, and generated the values of $x_2$. The results are presented in Fig. \ref{fig:ell2}. It clearly displays a stronger interference (or cross talk) between the elliptical models with respect to the values of $a$. The measure of this interference will help to determine parameters of the model for the protection of data privacy. We treat this interference as signal interference and apply blind signal separation approaches \citep{zarzoso1999blind} to determine the strength of data privacy.

\subsection{Blind Source Separation}
The blind source separation (BSS) is one of the classical techniques that is capable of separating the original signals from their copies of modulated signals without having any prior information about the original signals \citep{zarzoso1999blind}. The recent studies show that BSS is even capable of handling multidimensional data, like images and video (or image sequences) \citep{sorensen2013blind}. Therefore, we have adopted this technique as an attack approach \citep{liu2008survey} for the proposed perturbation model and derive robust parameters for the model. The standard measure used with BSS technique (or the attack) is called the Signal Interference Ratio (SIR), which is defined by the following fraction:
\begin{equation}
\rho = \frac{ps_m}{pc_t},
\label{eq:four}
\end{equation}
where $ps_m$ and $pc_t$ stand for the power of modulated signal and the power of cross-talk between the co-channels, respectively. The ratio $\rho$  is measured in decibel $dB$. When the denominator - power of cross-talk - increases, the ratio $\rho$ decreases, and it is hard to recover the source signals from the modulated signals. This fraction is defined based on the information available at \url{https://cran.r-project.org/web/packages/JADE/index.html}. It means that lower the SIR the higher the strength of modulation. The BSS technique states that if the SIR value is greater than 20 dB then the source signals ($x_1$ and $x_2$) are recoverable from $y_1$, and if the SIR values is less than or equal to 20 dB then source signals are not recoverable \citep{boscolo2004independent}, \citep{caiafa2005non}. We use this for the validation of proposed perturbation model.

\subsection{Random Forest Classification}
Among many classification techniques in a machine learning system, we have selected the random forest technique \citep{breiman2001random} for our research, because of its ability to address multi-class classification problem better than many other machine learning techniques, including support vector machine \citep{jeyakumar2014support,suthaharan2016support} and decision tree \citep{murthy1998automatic}. The random forest classifiers divide the data domain efficiently using bootstrapping technique - used to generate random decision trees - and Gini index - used to split the tree nodes. Hence it is highly suitable for the classification objectives of a large and imbalanced data set with many features.

\subsection{Misclassification and OOB Errors}
Several measures have been used to quantify the performance of classification techniques in machine learning; among them out-of-bag (OOB) error and misclassification errors are the most commonly used errors for the random forest classifiers \citep{breiman1996bagging}. OOB error is defined by the ratio between the total number of misclassified items from a set and the total number of items in the set. Similarly the misclassification error of a class is defined by the ratio between the number of misclassified items in the class and the total number of items in the class. We have used both of these quantitative measures to evaluate the performance of random forest classification algorithm in the input domain as well in the transform domain with the proposed perturbation model, and compare the results.

\section{Experimental Results}
We studied the performance degradation of random forest classifiers using the proposed elliptical perturbation model and the highly imbalanced NSL-KDD data set (\url{http://www.unb.ca/cic/research/datasets/nsl.html}), which we downloaded and used it in a previous research \citep{suthaharan2012relevance}. This data set has 25,192 observations with 41 network traffic features and 22 network traffic classes. We labeled the entire feature vector as ($f_1, f_2, \dots, f_{41}$), and reduced it later to a lower-dimensional feature vector, based on their importance to random forest classification. This data set forms the original domain and we represented this data set as ``dataset-O''. In this data set, the normal traffic class and the Neptune attack class have large number of observations, compared to other attack classes; hence, it provides a highly imbalanced data set. 

\begin{table}[h]
\caption{Statistical information of different traffic types in the NSL-KDD data set used - number of observations greater than or equal to 30.}
\vspace{3mm}
\centering
\begin{tabular}{|p{0.7cm}|p{1.6cm}|p{0.8cm}||p{0.7cm}|p{1.7cm}|p{0.7cm}|}
\hline
Label & Traffic & \#Obs. & Label & Traffic & \#Obs.\\
\hline
0 & Normal & 13449 & 11 & guess\_pwd & 10\\
\hline
1 & Neptune & 8282 & 12 & ftp\_write & 1\\
\hline
2 & back & 196 & 13 & multihop & 2 \\
\hline
3 & Warezclient & 181 & 14 & warezmaster & 7 \\
\hline
4 & ipsweep & 710 & 15 & loadmodule & 1 \\
\hline
5 & portsweep & 587  & 16 & spy & 1\\
\hline
6 & teardrop & 188 & 17 & imap & 5\\
\hline
7 & nmap & 301 & 18 & buf\_ovrflow & 6\\
\hline
8 & satan & 691 & 19 & land & 1\\
\hline
9 & smurf & 529 & 20 & phf & 2\\
\hline
10 & pod & 38  & 21 & rootkit & 4\\
\hline
\end{tabular}
\end{table}

The network traffic details of this data set presented in Table 1 clearly show the imbalanced nature of the data set between normal and attack traffic classes, and among the attack traffic classes. The first 11 traffic classes (labeled 0 to 10) presented in this table have more than 30 observations, and the next 11 traffic classes (labeled 11 to 21) have much less than 30 observations. One of the goals is to study the effect of the proposed perturbation model on the performance of random forest classifiers using the first 11 traffic classes only; however, we will use the other 11 traffic classes to understand imbalanced nature of the data and its significance to random forest classification. 

\subsection{Feature Selection using Random Forest}
There are 41 features - as we denoted by ($f_1, f_2, \dots, f_{41}$) earlier - in the dataset-O, and this feature vector determines the dimensionality 41 of the original domain; however, not necessarily all of these features contribute to the classification performance of random forest. To prepare the data set for our experiments and select the important features for classification, we first removed the categorical variables (or features) along with the features that overshadow the other features due to outliers. We then applied random forest classification to determine the importance of features by ordering them based on their misclassification errors. 

Using the approach suggested by \citep{zumel2014practical}, and by removing the least important feature from the feature vector one-by-one, while performing random forest classification repeatedly until a change in misclassification error can be observed. This process resulted in a lower-dimensional data set with 16 features, ($f_{33}$, $f_4$, $f_{32}$, $f_6$, $f_{36}$, $f_{20}$, $f_{28}$, $f_{19}$, $f_{31}$, $f_{27}$, $f_9$, $f_{29}$, $f_8$, $f_{23}$, $f_{37}$, $f_{30}$) in the decreasing order of importance. Hence, we have reduced the data set to a data set ($p=16$) with the most important feature vector that contributes to random forest classification. For simplicity, we represented these features by ($x_1, x_2, \dots, x_{16}$) respectively. Therefore, the dimension of the input domain of the proposed perturbation model is $p=16$ with 25,192 observations, 16 network traffic features, and 22 network traffic classes. Let's represent this dimension-reduced data set for the input domain as ``dataset-I''.

\subsection{Transform Domain Pattern Analysis}
The next step is to build the perturbation model, using the dataset-I as the input domain and construct the transform domain so that the random forest classifiers can be evaluated. Due to the pairing of features, multiple elliptical perturbation models were generated by selecting suitable parameters for the model, and they are discussed in the subsections below.

\subsubsection{Multiple Model Generation}
The proposed theoretical model for a single pair of features was presented in equation (\ref{eq:three}), which is applied to every consecutive pair of features: ($x_1, x_2$), $(x_3, x_4)$, \dots, ($x_{15}, x_{16}$) associated with the input domain; however, one can apply different techniques to select and combine the features. The pairing of these sixteen features of the input domain can give 8 models $M_i$ with new features $y_i$ for transform domain as follows:
\begin{equation}
y_i = \sqrt{a_i x_{2i-1}^2 + (1-a_i) x_{2i}^2} + \alpha \epsilon,
\label{eq:mult_model}
\end{equation}
where $i=1 \dots 8$; hence, we have 8 different models with elliptical patterns that form the transform domain with dimension 8. It is obvious that the parameters $a_i$, $i =1 \dots, 8$ together, and $\alpha$ contribute to the elliptical patterns and their distortion, and in turn contribute to the robustness of the proposed perturbation model to privacy attacks. They also contribute to the performance degradation of random forest classifiers in the transform domain. Therefore, a trade-off mechanism is required to achieve a strong privacy protection and a low misclassification error. The SIR measure is a flexible quantifier that allows a wide range of values to quantify the strength of privacy protection against BSS attack. The next subsection describes the empirical approach where we utilized this measure to find a set of values for the parameters $a_i$, $i =1 \dots, 8$ by fixing $\alpha = 0.001$.

\subsubsection{Parameter Selection for the Models}
We used Monte Carlo approach with the JADE implementation of SIR computation to assess BSS attack empirically. In this implementation, multiple copies of modulated source signals are generated using random weights, and then a SIR value is calculated to determine if the source signals are recoverable (if SIR is greater than 20dB then source signals are recoverable, otherwise they are not) from the multiple modulated signals. In our implementation, the feature pair ($x_{2i-1}$,$x_{2i}$), $i = 1, \dots, 8$ is considered as source signals, and $y_i$ is considered as their modulated signal. To create, multiple copies of modulated signal $y_i$, using ($x_{2i-1}$,$x_{2i}$), we generated several values for $a_i$ randomly from Uniform distribution, and used them in equation (\ref{eq:mult_model}). We then used the Monte Carlo approach to achieve desired results.

The Monte Carlo approach, combined with the JADE application of SIR and BSS attack provided us with the three values 0.042, 0.021, 0.096, which we selected for $a_1$, $a_2$, and $a_3$. To cut down the computational cost of Monte Carlo approach, we used them repeatedly for the parameters $a_i$ as follows: $a_1=0.042$, $a_2=0.021$, $a_3=0.096$, $a_4=0.042$, $a_5=0.021$, $a_6=0.096$, $a_7=0.042$, and $a_8=0.021$ for the 8 models, respectively. We obtained the SIR values for these parameters: 14.289, 10.983, 7.873, 11.483, 11.758, 12.608, 14.675, 16.235, respectively - the values less than 20dB indicate the source signal separation is difficult; hence, BSS attack is not possible. We can also see, each model has different privacy strengths, for example, model $M_3$ is much stronger than model $M_8$ against BSS attack. Therefore, in this step, we generated a data set for the transform domain, and it has 25,192 observations with 8 newly defined traffic features ($y_i$, $i=1, \dots, 8$) and 22 network traffic classes. Let's represent this transform domain data set as ``dataset-T''.

\subsection{Performance Degradation Evaluation}
We divided the performance degradation evaluation task into two experiments: ``experiment with full-imbalanced data sets'', and ``experiment with reduced-imbalanced data sets''. In the first experiment, we used the data sets dataset-I and dataset-T to compare the performance of random forest in both the input domain and transform domain. These two data sets have all 22 network traffic types with their full imbalanced traffic nature. As listed in Table 1, there are 11 traffic types with much fewer than 30 observations (totaling 40 observation) - the removal of these traffic types may influence the classification results. Hence, for the second experiment, we created two new data sets, dataset-IR and dataset-TR, from dataset-I and dataset-T, respectively. We removed the 40 observations related to these 11 traffic types. Hence the dataset-IR has 25,152 observations with dimension 16 and 22 traffic classes, and the dataset-TR has 25,152 observations with dimension 8 and 22 traffic classes.

\subsubsection{Experiment with full-imbalanced data sets:} We used both dataset-I and dataset-T to compare the performance of random forest classifiers in input domain and transform domain respectively. We conducted this experiment to evaluate the classification performance using random forest with the original (unprotected features) and transformed variables (protected features). The idea is to analyze the performance of random forest if the training is performed on these two full-imbalanced data sets. Therefore, we used both OOB error and misclassification error to compare the performances.

\paragraph{OOB error}The OOB errors and misclassification errors are presented in Tables 2 and 3 in their second and third columns, respectively. The tables also provide the information of the tuples, correctly classified and misclassified number of observations, for each class in input domain - denoted by ($idcc$, $idmc$) - and transform domain - denoted by ($tdcc$, $tdmc$), respectively. In the tables, the OOB errors are calculated as a single measure for the classification performance on the set, thus we have a single value of 0.0098 for input variables (unprotected features), 0.0169 for transformed variables (protected features). If we round these values to the second decimal places, we get 0.01 and 0.02 OOB errors, making it 1\% error difference in the performance degradation - input domain versus transform domain. We can see that the perturbation model increases the OOB error slightly while protecting data privacy.  

\paragraph{Misclassification error}Similarly, by comparing misclassification errors presented in Table 2 and Table 3, we observed that the perturbation model has a higher misclassification errors as expected, showing the characteristics of a perturbation model. As we can observe, the misclassification errors are increased, except for the traffic types ipsweep, teardrop, and pod. However, the error differences are significantly lower; hence, the perturbation model helps achieve both the protection of data privacy and the classification performance of random forest.

\begin{table}[t]
\caption{Input Domain: Random forest classification results of NSL-KDD data with original features and full-imbalanced data}
\vspace{3mm}
\centering
\begin{tabular}{|c|c|c|}         
\hline
Label & OOB errors & Misclassification errors\\
\hline
Normal & 0.0098 & 0.005 (13379, 70)\\
\hline
Neptune & 0.0098 & 0.003 (8256, 26)\\
\hline
back & 0.0098 & 0.025 (191, 5)\\
\hline
warezclient & 0.0098 & 0.127 (158, 23)\\
\hline
ipsweep & 0.0098 & 0.026 (691, 19)\\
\hline
portsweep & 0.0098 & 0.017 (577, 10)\\
\hline
teardrop & 0.0098 & 0.010 (186, 2)\\
\hline
nmap & 0.0098 & 0.086 (275, 26)\\
\hline
satan & 0.0098 & 0.041 (662, 29)\\
\hline
smurf & 0.0098 & 0.015 (521, 8)\\
\hline
pod & 0.0098 & 0.184 (31, 7)\\
\hline
\end{tabular}
\end{table}

\begin{table}[t]
\caption{Transform Domain: Random forest classification results of NSL-KDD data with EPA transformed features and full-imbalanced data}
\vspace{3mm}
\centering
\begin{tabular}{|c|c|c|}         
\hline
Label & OOB errors & Misclassification errors\\
\hline
Normal & 0.0169 & 0.009 (13322, 127)\\
\hline
Neptune & 0.0169 & 0.009 (8205, 77)\\
\hline
back & 0.0169 & 0.041 (188, 8)\\
\hline
warezclient & 0.0169 & 0.232 (139, 42)\\
\hline
ipsweep & 0.0169 & 0.021 (695, 15)\\
\hline
portsweep & 0.0169 & 0.063 (550, 37)\\
\hline
teardrop & 0.0169 & 0.005 (187, 1)\\
\hline
nmap & 0.0169 & 0.116 (266, 35)\\
\hline
satan & 0.0169 & 0.063 (647, 44)\\
\hline
smurf & 0.0169 & 0.045 (505, 24)\\
\hline
pod & 0.0169 & 0.053 (36, 2)\\
\hline
\end{tabular}
\end{table}

\begin{table}[t]
\caption{Input Domain: Random forest classification results of NSL-KDD data with original features and reduced-imbalanced data}
\vspace{3mm}
\centering
\begin{tabular}{|c|c|c|}         
\hline
Label & OOB errors & Misclassification errors\\
\hline
Normal & 0.0088 & 0.005 (13381, 68)\\
\hline
Neptune & 0.0088 & 0.003 (8253, 29)\\
\hline
back & 0.0088 & 0.025 (191, 5)\\
\hline
warezclient & 0.0088 & 0.127 (158, 23)\\
\hline
ipsweep & 0.0088 & 0.025 (692, 18)\\
\hline
portsweep & 0.0088 & 0.013 (579, 8)\\
\hline
teardrop & 0.0088 & 0.010 (186, 2)\\
\hline
nmap & 0.0088 & 0.093 (273, 28)\\
\hline
satan & 0.0088 & 0.044 (660, 31)\\
\hline
smurf & 0.0088 & 0.015 (521, 8)\\
\hline
pod & 0.0088 & 0.210 (30, 8)\\
\hline
\end{tabular}
\end{table}

\begin{table}[t]
\caption{Transform Domain: Random forest classification results of NSL-KDD data with EPA transformed features and reduced-imbalanced data}
\vspace{3mm}
\centering
\begin{tabular}{|c|c|c|}         
\hline
Label & OOB errors & Misclassification errors\\
\hline
Normal & 0.0156 & 0.009 (13322, 127)\\
\hline
Neptune & 0.0156 & 0.009 (8207, 75)\\
\hline
back & 0.0156 & 0.040 (188, 8)\\
\hline
warezclient & 0.0156 & 0.220 (141, 40)\\
\hline
ipsweep & 0.0156 & 0.022 (694, 16)\\
\hline
portsweep & 0.0156 & 0.061 (551, 36)\\
\hline
teardrop & 0.0156 & 0.005 (187, 1)\\
\hline
nmap & 0.0156 & 0.102 (270, 31)\\
\hline
satan & 0.0156 & 0.059 (650, 41)\\
\hline
smurf & 0.0156 & 0.039 (508, 21)\\
\hline
pod & 0.0156 & 0.053 (36, 2)\\
\hline
\end{tabular}
\end{table}

\subsubsection{Experiment with reduced-imbalanced data sets} We used dataset-IR and dataset-TR to compare the performance of random forest classifiers in input and transform domains for the purpose of this experiment. It means only the 11 traffic types with more than 30 observations were classified to study if there was any significant effect due to the elimination of other traffic types that have significantly lower number of observations. The results are presented in Tables 4 and 5, and we can observe similar patterns between the input domain and transform domain results. Hence, comparing the results in Tables 2 and 4, we can see that the OOB error has slightly decreased due to the reduced-imbalanced nature of traffic types, as expected. Similarly, comparing the results in Tables 3 and 5, we can see the reduction in the OOB error, and an overall reduction in the misclassification errors.

\subsection{Overall Performance Degradation}
Although, the results presented in the previous section provide information to compare the performance degradation of the random forest classifiers between the input domain and the transform domain, it is important to understand the overall performance degradation to conclude if the proposed perturbation is meaningful. Therefore, to estimate the percentage performance degradation, we defined a simple measure:
\begin{equation}
pd_t = \frac{tdmc_t - idmc_t}{tot_t}.
\label{eq:deg}
\end{equation}
For example, the transform domain misclassification ($tdmc_t$) of traffic type ``normal'' is 127 (from Table 3), and the input domain misclassification ($idmc_t$) of traffic type ``normal'' is 70 (from Table 2). Also the total number of observations of ``normal'' traffic class is 13449 (Table 1). Therefore, the percentage degradation of random forest by the proposed perturbation model for the ``normal'' class is 0.4238233. Similarly, we calculated the percentage degradations for other 10 traffic types with full-imbalanced data sets, and listed all of them in Table 6 (column 2). We also calculated the same for reduced-imbalanced data sets, and provided the results in column 3 of Table 6. \textit{Note that a positive value indicates it is a degradation over input domain to transform domain, whereas, a negative value indicates there is an improvement over input domain to transform domain}. The average degradations over all the class types are 1.05\% for full-imbalanced data sets, and 0.45\% for reduced-imbalanced data sets - indicating additional average degradation of 1.05\% when the full-imbalanced data is used, additional average degradation of 0.45\% when reduced-imbalanced data is used, and the difference shows the use of additional imbalanced data affects the performance negatively.  

\begin{table}[t]
\caption{Performance degradation of random forest classifiers over input domain to EPA transformed domain using full/reduce-imbalanced data}
\vspace{3mm}
\centering
\begin{tabular}{|c|c|c|}         
\hline
Label (t) & Full-Imb. ($pd_t$) & Reduced-Imb. ($pd_t$)\\
\hline
Normal & 0.4238233 & 0.4386943\\
\hline
Neptune & 0.6157933 & 0.5554214\\
\hline
back & 1.5306122 & 1.5306122\\
\hline
warezclient & 10.4972376 & 9.3922652\\
\hline
ipsweep & -0.5633803 & -0.2816901\\
\hline
portsweep & 4.5996593 & 4.7700170\\
\hline
teardrop & -0.5319149 & -0.5319149\\
\hline
nmap & 2.9900332 & 0.9966777\\
\hline
satan &  2.1707670 & 1.4471780\\
\hline
smurf & 3.0245747 & 2.4574669\\
\hline
pod & -13.1578947 & -15.7894737\\
\hline
\hline
AVG. ERR. & 1.054483 & 0.4532049\\
\hline
\end{tabular}
\end{table}

\section{Competing Methods and Discussion}
We have selected PCA as the competing method to evaluate the performance of the proposed EPA approach. PCA is a classical linear transformation which transforms the original features to principal components (PCs), hence achieves effective dimension reduction \citep{du2014principal}. It has been extensively used in modern applications, including atmospheric science \citep{jolliffe2016principal}, neuroscience \citep{lee2016selective}, and neuroimaging \citep{jones2007novel}. It became popular in the last two decades because of the recent developments in computer technology that can help the application of PCA to high dimensional large data sets. However, it generally suffers from two major drawbacks as reported in \citep{bruce2017practical}. One of them is the strong statistical assumptions and the second one is the difficulty of selecting the number of PCs for dimensionality reduction and achieve data utility. 

\subsection{Comparative Analysis}
The results of PCA transformation - applied to the full-imbalanced NSL-KDD data - are presented in Table 7 and they can be compared with the results of the proposed EPA approach (applied to the same data) in the second column of Table 6. We adopted two criterion to extract number of PCs: eigenvalue greater than 1 criterion (i.e., Kaiser-Guttman criterion) as used in \citep{hung2016development} and 80\% cumulative variance rule as stated in \citep{bruce2017practical}. The number of PCs selected by these criterion are 5 and 6, respectively. The random forest classification results ($pd_t$) using the first 5 PCs and 6 PCs of this data are presented in the second and third columns of Table 7.

\subsubsection{General Analysis}
The results in the second columns of Tables 6 and 7 show that the average performance degradation caused by PCA with 5 PCs is higher (almost double) than the degradation caused by the proposed EPA approach. In contrast, the results in the third column suggests a smaller degradation is possible if 6 PCs are used. These results, with the use of higher number of PCs, PCA can achieve better classification accuracy; however, it also suggests the proposed approach can be competitive too.

\subsubsection{Specific Analysis}
In network security, Denial-of-Service (DoS) attack is generally considered a major threat to network users and the servers. Therefore, the classification of Normal traffic and DoS attacks are very important. The DoS attack includes the attacks such as Neptune, Back, Teardrop, Smurf and Pod \citep{jin2007network} and they are included in NSL-KDD data set as well. Therefore, we calculated the performance degradation ($pd_t$) for these attacks separately and obtained  -1.35, 1.97, and 0.67 for EPA, PCA with 5 PCs, and PCA with 6PCs, respectively. The negative value, as stated earlier, indicates an improvement in the performance; thus, It shows the proposed EPA is superior than PCA when the classification of DoS attacks are considered.
 
In terms of invertible characteristics, according to \citep{geiger2014information}, it is possible to invert PCA with an estimate of the covariance matrix; hence, it is relatively weaker than the proposed EPA approach when the applications such as data privacy and security are considered. However, in terms of dimension reduction, PCA can be superior than the proposed method because it can reduce the dimension by more than 50\%, whereas the proposed EPA approach has the fixed 50\% dimension reduction.

\begin{table}[t]
\caption{Performance degradation of random forest classifiers over input domain to PCA transformed domain using full-imbalanced data only}
\vspace{3mm}
\centering
\begin{tabular}{|c|c|c|}         
\hline
Label (t) & Full-Imb. 5PC ($pd_t$) & Full-Imb. 6PC ($pd_t$)\\
\hline
Normal & 0.3345974 & 0.1487099\\
\hline
Neptune & 0.4829751 & 0.4346776\\
\hline
back & 13.7755102 & 10.7142857\\
\hline
warezclient & 7.1823204 & 3.3149171\\
\hline
ipsweep & 0.4225352 & 0.7042254\\
\hline
portsweep & 1.8739353& 1.7035775\\
\hline
teardrop & -0.5319149 & -0.5319149\\
\hline
nmap & 0.9966777 & 0.3322259\\
\hline
satan &  0.8683068 & 0.5788712 \\
\hline
smurf & 5.6710775 & 6.4272212\\
\hline
pod & -7.8947368 & -13.1578947\\
\hline
\hline
AVG. ERR. & 2.107389 & 0.9699002\\
\hline
\end{tabular}
\end{table}

\subsection{Evaluation using IRIS plant data set}
We also used the iris plant dataset to evaluate and compare EPA and PCA transformations. This dataset is a simple, yet effective dataset, which has been used in machine learning extensively for the last several decades \citep{chaudhary2016hybrid, timon2016parallel, lin2017relative}. We obtained this data from the UCI Machine Learning Repository \citep{Lichman:2013}. Random forest is applied to the original iris data, OOB errors are calculated and presented in the second column of Table 8. The data is then transformed into PCs using PCA. The random forest classification is applied using all the PCs and the OOB results are presented in the third column of Table 8. We also transformed the data set using the proposed EPA transformation and the applied random forest classification. The OOB results of the proposed approach is presented in the fourth column of the table. Note that the first column of the table shows the three classes of the iris plant. Comparing the results in Table 8, we can say that the proposed transformation provides the classification results closer to the results of random forest applied to the original data than the principal components.

\section{Conclusion}
This study allowed us to understand the variations caused by the perturbation models between their input domain and transform domain characteristics or numerical patterns. This knowledge helped us construct a parametric perturbation model using an elliptical transformation along with an additive Gaussian noise degradation. The degradation performance analysis using random forest classifiers together with blind source separation attack and quantitative measures - signal interference ratio, OOB error, and misclassification error - showed that the parametric elliptical perturbation model performed very well in the classification of network intrusion and biological data, while protecting data privacy patterns of feature vectors of the data.

Compared with classical linear transformations such as PCA,  the proposed method requires less statistical assumptions on the data and is highly suitable for the applications such as data privacy and security as a result of the difficulty of inverting the elliptical patterns from the transform domain to the input domain. In addition, we adopted a flexible block-wise dimension reduction step in the proposed method to accommodate the possible high-dimensional data ($p \gg n$) in modern applications, in which PCA is not directly applicable. The empirical performance results also confirmed the superior performance of the proposed EPA approach over the widely used PCA. 

Several future directions still remain of interest in our research agenda. First, the current paper mainly discusses pairing of two features (block size is 2) and fixed projections. It is possible to consider larger block sizes and random projections to reduce computation complexity. Second, model (3) can be extended by replacing the constraint $a+b=1$ with flexible alternatives, and by considering a diagonal elliptical models.

\begin{table}[t]
\caption{OOB errors of three cases using IRIS plant data set}
\vspace{3mm}
\centering
\begin{tabular}{|p{1.3cm}|c|c|c|}
\hline
Class & OOB: RF & OOB: RF-PCA & OOB: RF-EPA\\
\hline
Setosa & 0.00 & 0.00 & 0.00\\
\hline
Versicolor & 0.08 & 0.12 & 0.12\\
\hline
Virginica & 0.06 & 0.10 & 0.06\\
\hline
\end{tabular}
\end{table}

\section*{Acknowledgments}
This research of the first author was partially supported by the Department of Statistics, University of California at Irvine. 

\bibliographystyle{model2-names}
\bibliography{refs}

\end{document}